\documentclass[10pt,twocolumn,letterpaper]{article}

\usepackage[pagenumbers]{wacv}

\definecolor{wacvblue}{rgb}{0.21,0.49,0.74}
\usepackage[pagebackref,breaklinks,colorlinks,allcolors=wacvblue]{hyperref}
\usepackage{makecell} 
\usepackage{multirow}
\usepackage{algorithm}
\usepackage{algpseudocode}
\usepackage[utf8]{inputenc}
\usepackage{newunicodechar}
\def\wacvPaperID{279} %
\def\confName{WACV}
\def\confYear{2027}

\title{FastVLA: Task-Conditioned and Temporally Adaptive Visual Token Compression for Efficient Vision-Language-Action Models}

\author{Jiuyi Xu\\
Colorado School of Mines\\
\and
Chengyuan Qian\\
Texas A\&M University\\
\and
Zhengzhong Tu\\
Texas A\&M University\\
\and
Yang Sui\\
Microsoft AI\\
\and
Yangming Shi\\
Colorado School of Mines\\
}

\begin{document}
\maketitle
\begin{abstract}

Vision-Language-Action (VLA) models present a promising paradigm for general-purpose robot manipulation, but closed-loop deployment is often limited by inference efficiency: at each timestep, high-resolution, often multi-view images are encoded into hundreds of visual tokens that dominate transformer computation and latency. 
Token compression via attention is a natural solution but is challenging for VLA control because (1) decision-critical tokens are determined by the task, not generic visual saliency, and (2) the token budget varies substantially across steps, so static metrics and budgets waste computation or discard control-critical tokens. 
We introduce FastVLA, a training-free, task-aware, and temporally adaptive visual token compression framework. 
At each step, FastVLA scores token importance via task-to-vision attention and adaptively sets the compression rate from the importance-distribution uniformity, measured by an inverse-Simpson metric with online sliding-window calibration. 
We evaluate FastVLA on OpenVLA, OpenVLA-OFT, and CogACT across LIBERO and SIMPLER, and on a physical Franka Research 3 robot. 
In simulation, it reduces LLM-backbone FLOPs by 35-59\% and action latency by 9-32\% (a 1.09$\times$-1.46$\times$ speedup) while maintaining task success, achieving a comparable or better efficiency-performance trade-off than existing plug-and-play token compression/pruning/reuse baselines. 
On the real robot, it cuts action-prediction latency by 20\% (1.25$\times$ speedup) at comparable success. 
Code is available at \url{https://github.com/jiuyixu25/FastVLA}.

\end{abstract}
    
\section{Introduction}
Vision-Language-Action (VLA) models have recently emerged as a compelling paradigm for general-purpose robot manipulation, enabling policies to predict low-level actions conditioned on natural-language tasks and visual observations~\cite{zitkovich2023rt,kim2024openvla,black2024pi_0,team2025gemini,cai2026xiaomi}.
By leveraging large-scale pretraining and large vision-language backbones, VLAs demonstrate strong generalization across objects, scenes, and task specifications, making them increasingly attractive for deployment in versatile manipulation systems.
Despite their impressive capabilities, deploying these models in closed-loop control is often limited by inference efficiency~\cite{chi2025diffusion}.
At every control step, high-resolution (e.g., 224$\times$224) and frequently multi-view observations are encoded into hundreds of visual tokens~\cite{kim2025fine,intelligence2025pi_} that dominate the compute and latency of the vision-language-model (VLM) backbone during action prediction, far outweighing the few text tokens of a short task.

Yet not all visual tokens contribute equally to action prediction.
In a typical manipulation step, only a small subset of tokens carries decision-critical information (e.g., the target object/place or the gripper--object contact region), while the rest are redundant or even distracting~\cite{xu2025vla,yang2025efficientvla,jiang2025better}.
Reducing the number of visual tokens processed by the large-language-model (LLM) backbone, without sacrificing task performance, has therefore become a key direction for accelerating VLA inference~\cite{pei2025action,tan2025think,wang2025specprune}.

However, directly applying VLM-specific token compression methods~\cite{bolya2022token,chen2024image,zhang2024sparsevlm} to robotic control is often suboptimal, because these methods are designed for single-shot vision-language understanding, where the objective is to preserve global semantic content for perception tasks such as detection, captioning, or generation.
Unlike image-based VLMs that process a single image, or video-based VLMs that encode a fixed sequence of frames, a VLA interacts with the environment through actions that directly shape its subsequent observations: at each step the agent receives a new observation conditioned on its previous action and predicts the next one.
Visual tokens are thus not representations of a static scene but components of a \emph{sequential decision process}, which sets token compression for VLA apart from its VLM counterpart in two respects.
The first is that token relevance is governed by the task rather than by generic visual saliency (now standard practice in language-guided pruning), which we adopt as our importance signal rather than claim as a contribution.
The second, and the focus of this work, is that the \emph{amount} of decision-critical visual information varies drastically across control steps: some steps (e.g., target localization) rely on broad scene context, whereas others (e.g., grasping or contact) depend on a small, localized region.
A single fixed compression rate is therefore inherently mismatched, wasting computation when the useful information is concentrated, and discarding necessary context when broader information is required.

Motivated by this, we propose \textbf{FastVLA, a training-free visual token compression framework for efficient VLA inference whose core is a temporally adaptive, per-step compression rate}.
At each step, FastVLA scores visual-token importance with task-to-vision attention which is a training-free signal that a controlled comparison shows is more reliable for action prediction than vision-to-vision or all-to-vision attention and then, crucially, decides \emph{how many} tokens to keep from the shape of the resulting importance distribution.
Specifically, it estimates the effective number of tokens the decision depends on via an inverse-Simpson uniformity measure, and maps this quantity through an online sliding-window calibration to a timestep-dependent compression rate: when the current step concentrates on few tokens relative to its recent history it prunes aggressively, and when the decision draws on broader context it prunes conservatively.
By allocating the token budget where each decision needs it, FastVLA reduces LLM-backbone computation, speeds up action prediction, and improves closed-loop responsiveness, without caches or training.

We evaluate FastVLA on three open-source VLA models including OpenVLA~\cite{kim2024openvla}, OpenVLA-OFT~\cite{kim2025fine}, and CogACT~\cite{li2024cogact}, across the widely used LIBERO~\cite{liu2023libero} and SIMPLER~\cite{li24simpler} benchmarks, and on a physical Franka Research 3 robot.
In simulation, FastVLA reduces LLM-backbone FLOPs by 35-59\% and action latency by 9-32\% (up to a $1.46\times$ speedup) while maintaining task success, yielding a comparable or even better efficiency-performance trade-off than existing training-free compression, pruning, and reuse baselines.
On the real robot, it cuts action-prediction latency by 20\% ($1.25\times$ speedup) and raises the control frequency by up to 25\% at comparable success.
In summary, our contributions are:
\begin{itemize}
  \item We propose FastVLA, a training-free, plug-and-play visual token compression framework that accelerates closed-loop VLA inference.
  \item We design a step-adaptive compression policy that sets a timestep-dependent token budget from the inverse-Simpson uniformity of the importance distribution under online sliding-window calibration.
  \item We validate FastVLA on OpenVLA/OpenVLA-OFT (LIBERO), CogACT (SIMPLER), and a physical Franka Research 3 robot, achieving substantial compute and latency reductions in simulation and a $1.25\times$ real-robot speedup with comparable task success.
\end{itemize}

\begin{figure*}
    \centering
    \includegraphics[width=1\linewidth]{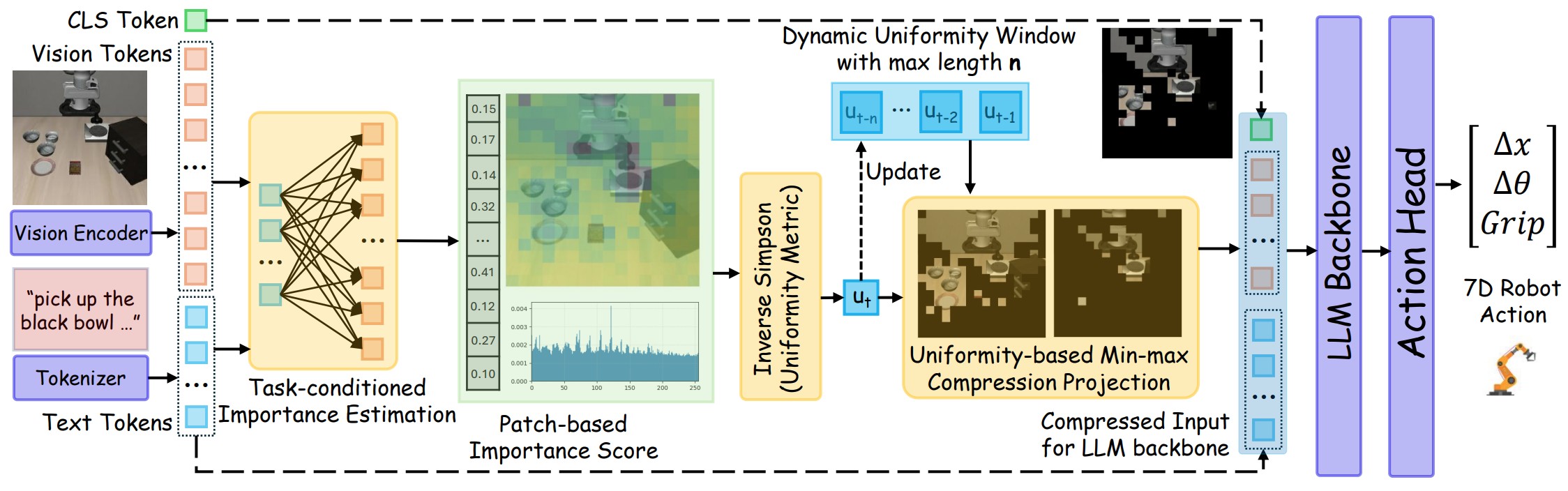}
    \caption{Overview of FastVLA. At each action-prediction step, FastVLA scores visual tokens with task-conditioned attention, measures the uniformity of the top-scoring importance distribution, and sets the step's compression rate through an online sliding-window calibration, reducing LLM-backbone computation while preserving action performance.}
    \label{fig:framework}
    \vspace{-10pt}
\end{figure*}

\section{Related Work}
\label{sec:related}

\subsection{Vision-Language-Action Models for Robot Manipulation}
VLA models predict robot actions directly from natural-language tasks and visual observations.
Early work such as SayCan~\cite{ahn2022can} combines language-conditioned planning with learned low-level skills, while RT-1~\cite{brohan2022rt} and RT-2~\cite{zitkovich2023rt} scale language-conditioned visuomotor learning with large robot datasets and web-scale vision-language pretraining.
In parallel, PaLM-E~\cite{driess2023palm} and Octo~\cite{team2024octo} highlight the benefits of large pretrained representations across tasks and embodiments.
More recently, foundation VLA models including OpenVLA~\cite{kim2024openvla} and OpenVLA-OFT~\cite{kim2025fine}, CogACT~\cite{li2024cogact}, and $\pi_0$~\cite{black2024pi_0} / $\pi_{0.5}$~\cite{intelligence2025pi_} demonstrate strong robustness and open-world generalization via heterogeneous multi-robot data and transformer backbones.
These billion-scale models encode high-resolution, often multi-view observations into long visual-token sequences at every control step, which motivates inference-time methods that remove redundant visual tokens while preserving task-critical information.

\subsection{Visual Token Compression for Efficient VLA Inference}
Because most VLA models~\cite{kim2024openvla,kim2025fine,li2024cogact,black2024pi_0,intelligence2025pi_} adopt VLM-style architectures, training-free token compressors built for VLMs are broadly applicable: they prune or merge visual tokens by token similarity (ToMe~\cite{bolya2022token}), shallow-layer or cross-modal attention (FastV~\cite{chen2024image}, SparseVLM~\cite{zhang2024sparsevlm}, VisPruner~\cite{zhang2025beyond}), or token diversity (DivPrune~\cite{alvar2025divprune}). However, these methods are designed for single-image inference and apply a static pruning policy, so transferring them directly to VLA often degrades control performance.

Recent work therefore targets VLA inference along three lines: reuse- and cache-based methods that exploit cross-step temporal redundancy (VLA-Cache~\cite{xu2025vla}, FlashVLA~\cite{tan2025think}); importance-estimation methods that add task-, phase-, or motion-aware cues (SpecPrune-VLA~\cite{wang2025specprune}, ADP~\cite{pei2025action}, VLA-Pruner~\cite{liu2025vla}); and hybrid or learned schemes that go beyond pruning (EfficientVLA~\cite{yang2025efficientvla}, LightVLA~\cite{jiang2025better}).

These methods still share limitations that FastVLA overcomes. Reuse and caching introduce cross-step state and cache-management overhead, so wall-clock gains can lag FLOP reductions; learned schemes such as LightVLA require training and are tied to a specific backbone, although at equal success they can reach a better compute–latency point than training-free methods. In addition, some VLM/VLA compression algorithms commit to a fixed compression ratio, adapting (if at all) only through auxiliary external signals such as end-effector motion. 
FastVLA differs on three axes: it is training-free, pruning independently at each step without caches; it conditions token importance on the language task; and, crucially, it makes the compression rate itself adaptive, reading the per-step concentration of task-relevant evidence via an inverse-Simpson uniformity signal calibrated against a short temporal window, spending tokens where the decision needs them and pruning harder where it does not.

\section{Method}
We propose FastVLA, a temporally adaptive and training-free visual token compression method that reduces the number of visual tokens processed by the VLA model's LLM backbone (Fig.~\ref{fig:framework}).
FastVLA proceeds in two stages: it first scores visual-token importance with a task-conditioned attention signal, and then, as our core contribution, sets \emph{how many} tokens to keep adaptively at each step from the concentration of that importance distribution.
We present both stages below.

\subsection{Preliminary: Visual Token Compression in VLA Models}
Methods that reduce the visual-token cost of a VLA backbone fall into three largely orthogonal families: \emph{token pruning} discards a subset of visual tokens, shortening the sequence seen by later layers; \emph{token merging} pools similar tokens into fewer representatives; and \emph{reuse/caching} recomputes only the tokens that changed across control steps, reusing cached token or key-value states for the rest.
FastVLA belongs to the pruning family: at every action step it performs a hard, task-conditioned top-$K$ selection of visual tokens, with the retained budget set adaptively per step.

\subsection{Task-conditioned visual token importance}
A VLA policy is explicitly conditioned on the language task, so visual-token usage is driven by the task prompt rather than by generic visual saliency: the tokens that matter are those covering the target object/place and the interaction region named by the task.
The design question is therefore \emph{which} attention signal best captures this task-driven importance.
Since selecting tokens by attention is already standard in VLM compression, we do not propose it as new but instead re-evaluate three candidate signals for VLA control: \emph{all-to-vision} (naive cross-modal attention from the fused representation to visual tokens), \emph{task-to-vision} (task tokens attending to visual tokens), and \emph{vision-to-vision} (visual self-attention capturing perceptual saliency).

\begin{figure}[t]
  \centering
  \includegraphics[scale=0.75]{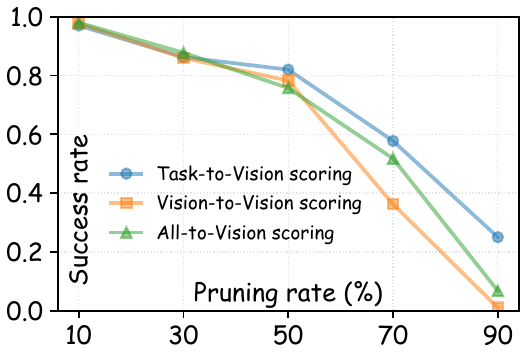}
  \caption{Average success rate on LIBERO under increasing visual token pruning rates for the three attention-based importance signals.}
  \label{fig:importance}
  \vspace{-10pt}
\end{figure}

We progressively pruned visual tokens by each score and measured policy performance on LIBERO with OpenVLA-OFT (Fig.~\ref{fig:importance}). Task-to-vision scoring was clearly the most robust: it retained near-saturated success under light-to-moderate compression (10\%--50\%) and preserved non-trivial performance even under aggressive compression ($\ge 70\%$), whereas vision-to-vision and all-to-vision degraded much faster and approached failure at 90\%. We therefore adopt task-to-vision attention as the importance signal in the remainder of the method.

Formally, let $\mathbf{T} \in \mathbb{R}^{M \times d}$ and $\mathbf{V} \in \mathbb{R}^{N \times d}$ denote the task- and visual-token representations. Following scaled dot-product attention~\cite{vaswani2017attention}, we apply RMSNorm to obtain $\tilde{\mathbf{T}}$ and $\tilde{\mathbf{V}}$, compute affinities $\mathbf{S} = \tilde{\mathbf{T}}\tilde{\mathbf{V}}^\mathsf{T}/\sqrt{d} \in \mathbb{R}^{M \times N}$, and apply a $softmax$ over visual tokens to obtain attention weights $\mathbf{A}$. Averaging over a subset of task tokens $\mathcal{Q}$ gives a per-token importance score:
$$
s_i = \frac{1}{|\mathcal{Q}|}\sum_{q \in \mathcal{Q}}\mathbf{A}_{q,i}, \quad i = 1,\dots,N,
$$
so that visual tokens consistently attended by the task context receive higher importance.

\subsection{Adaptive Step-aware Compression Rate}
\begin{figure}[t]
    \centering
    \includegraphics[width=1\linewidth]{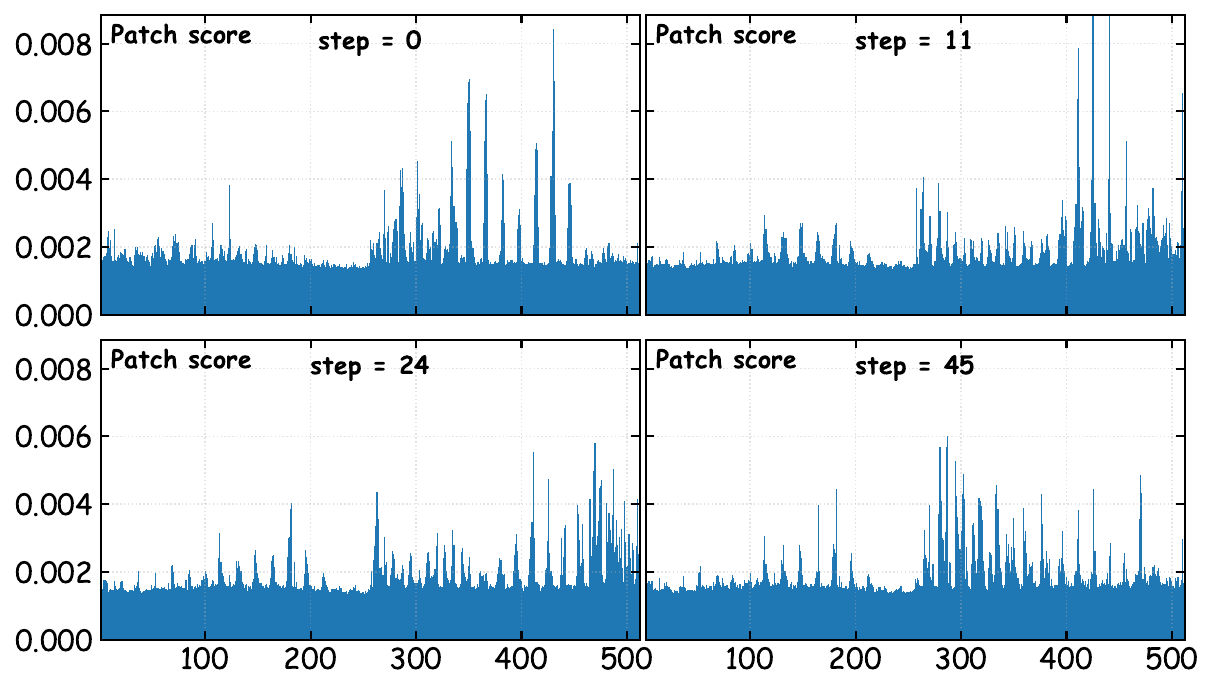}
    \caption{Visual token importance distributions across action steps in a single trial on the LIBERO-Long benchmark.}
    \label{fig:step}
\end{figure}

Having fixed \emph{which} tokens matter, the central question of our method is \emph{how many} to keep at each step.
In a closed-loop rollout, successive steps rely on visual information to very different extents. For example, approaching or grasping an object depends on a small localized region, whereas locating a target or verifying the scene requires broad context.
Fig.~\ref{fig:step} makes this concrete: within a single trial, the per-step importance distribution swings between nearly uniform (many tokens contribute comparably) and highly concentrated (a few tokens dominate).
A fixed compression rate therefore wastes computation on concentrated steps and discards needed context on broad ones; the rate should instead adapt to this per-step concentration.

\begin{figure}[t]
  \centering
  \includegraphics[width=0.75\linewidth]{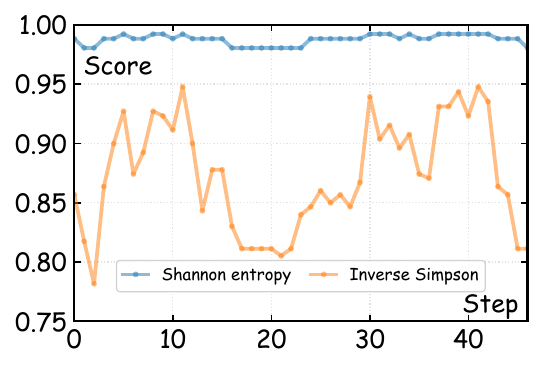}
  \caption{Shannon entropy and inverse Simpson over the action steps of a single LIBERO-Long trial.}
  \label{fig:entropy_vs_simpson}
  \vspace{-10pt}
\end{figure}

We implement this in two steps: measure how concentrated the importance distribution is, then map that concentration to a per-step rate.
To measure it, a natural choice is Shannon entropy, but with hundreds of visual tokens and a long tail of near-zero scores, normalized entropy varies only within a narrow band and is insensitive to concentration among the dominant tokens (Fig.~\ref{fig:entropy_vs_simpson}), making it hard to map to distinct compression rates.

\textbf{Inverse Simpson uniformity.} We instead adopt the second-order effective-number measure (inverse Simpson)~\cite{jost2006entropy,keylock2005simpson}, which emphasizes squared weights and thus down-weights the long tail while reporting the \emph{effective number} of dominant tokens which is a better fit for the sparse, heavy-tailed importance distributions of VLA models, and one with a substantially larger dynamic range across steps.
Over the top-$K$ tokens we compute:
\[
\begin{gathered}
\mathbf{v}=\mathrm{TopK}(\{s_i\},K),\qquad
q_j = \frac{v_j}{\sum_{m=1}^K v_m},\\[4pt]
K_{\text{eff}}^{(t)} = \frac{1}{\sum_{j=1}^K q_j^2},\qquad
U^{(t)} = \frac{K_{\text{eff}}^{(t)}}{K},
\end{gathered}
\]
where $U^{(t)} \in (0,1]$: small values indicate concentrated attention (few tokens dominate), large values a more uniform distribution. Here $K=\lceil(1-(\rho+\tau))N\rceil$ is the smallest budget the band can
request; since $K$ is fixed within a run and $U^{(t)}$ enters the decision only
through the calibration below, which is invariant to constant rescaling, this
normalization cannot conflate concentration with budget.

\begin{algorithm}[t]
\caption{FastVLA}
\label{algo}
\begin{algorithmic}[1]
\Statex \textbf{Input:} Current timestep $t$, task tokens $\mathbf{T}^{(t)}\in\mathbb{R}^{M\times d}$, visual tokens $\mathbf{V}^{(t)}\in\mathbb{R}^{N\times d}$, sliding window (queue) $\mathcal{W}$ with max length $T$
\Statex \textbf{Hyperparameters:} Target compression rate $\rho$, deviation factor $\tau$
    \State \textit{// Step 1: Attention-based Token Importance Score}
    \State $\tilde{\mathbf{T}} \gets \textsc{RMSNorm}(\mathbf{T}^{(t)})$, $\tilde{\mathbf{V}} \gets \textsc{RMSNorm}(\mathbf{V}^{(t)})$
    \State $\mathbf{s} \gets \textsc{Mean}\!\left(\textsc{Softmax}\!\left(\frac{1}{\sqrt{d}} \tilde{\mathbf{T}}\tilde{\mathbf{V}}^\top\right)\right)$
    \State \textit{// Step 2: Inverse Simpson-based Uniformity}
    \State $\mathbf{v} \gets \textsc{TopK}(\mathbf{s}, \left\lceil(1 - (\rho + \tau))N \right\rceil)$
    \State $\mathbf{q} \gets \mathbf{v}/\sum_j v_j;\ 
U^{(t)} \gets \frac{1}{\left\lceil(1 - (\rho+\tau))N \right\rceil \sum_j q_j^2}$
    \State \textit{// Step 3: Online Mapping based on Sliding Window}
    \If{$t < 2$} // \textit{Initialize $U_{\min}$ and $U_{\max}$}
        \State $\rho^{(t)} \gets \rho - \tau$
    \Else
        \State $U_{\min} \gets \textsc{Min}(\mathcal{W})$, $U_{\max} \gets\textsc{Max}(\mathcal{W})$
            \State $\alpha^{(t)} \gets \textsc{Clip}\!\left(\frac{U^{(t)} - U_{\min}}{U_{\max} - U_{\min}}, 0, 1\right)$
            \State $\rho^{(t)} \gets \textsc{Project}(\alpha^{(t)}; [\rho - \tau, \rho + \tau])$
    \EndIf
    \State $\mathcal{W} \gets \textsc{PushPop}(\mathcal{W}, U^{(t)}, T)$ 
\State \textbf{Return:} $\rho^{(t)}$
\end{algorithmic}
\end{algorithm}

\textbf{Online step-adaptive rate.} Thresholding $U^{(t)}$ against an absolute
scale would require anchors that do not transfer: the range of $U^{(t)}$ drifts
across suites and backbones, so anchors fit on one setting misallocate the
budget on another. We therefore calibrate $U^{(t)}$ against its own recent
temporal context through a sliding window $\mathcal{W}^{(t)} =
\{U^{(t-T)},\dots,U^{(t-1)}\}$ of length $T$. Letting $U_{\min}^{(t)}$ and $U_{\max}^{(t)}$ be the window extremes, the relative position
\[
\alpha^{(t)} = \mathrm{clip}\!\left(\frac{U^{(t)} - U_{\min}^{(t)}}{U_{\max}^{(t)} - U_{\min}^{(t)}},\,0,\,1\right)
\]
locates the current step within the ongoing execution: a small $\alpha^{(t)}$ (more concentrated than the recent past) warrants aggressive pruning, a large $\alpha^{(t)}$ (more uniform) warrants caution. We then map it to a per-step rate via a bounded linear projection centered at a target rate $\rho$:
\[
\rho^{(t)} = \mathrm{clip}\big(\rho + \tau\,(1-2\alpha^{(t)}),\;\rho-\tau,\;\rho+\tau\big),
\]
so the rate adapts within $[\rho-\tau, \rho+\tau]$, where $\rho$ sets the average compression (e.g., $50\%$) and the deviation factor $\tau$ bounds the per-step deviation (e.g., $\pm 10\%$). Algorithm~\ref{algo} summarizes the full per-step procedure.

\noindent\textbf{Implementation details.} Scores are computed on the LLM backbone's input embeddings with a parameter-free RMSNorm. $\mathcal{Q}$ contains every token after the visual span (the prompt with the task), skipping all-zero placeholder embeddings such as OpenVLA-OFT's action slots. $\mathcal{W}$ is cleared at the start of every episode. The first two steps use the conservative rate $\rho-\tau$ because $\mathcal{W}$ cannot yet provide a range, and a degenerate window ($U_{\max}=U_{\min}$) falls back to $\rho$. Since $\alpha^{(t)}$ is clipped, a step more concentrated (more uniform) than every step in $\mathcal{W}$ saturates at $\rho+\tau$ ($\rho-\tau$). Each step keeps the $\mathrm{round}((1-\rho^{(t)})N)$ top-scoring visual tokens in their original order.

\begin{table*}[!t]
    \caption{Comparison of different VLA acceleration methods on the LIBERO benchmark. 
    FastVLA uses $\rho{=}0.8/0.7/0.75/0.6$ on OpenVLA-OFT and $0.35/0.5/0.4/0.3$ on OpenVLA (Spatial/Object/Goal/10).
    FastVLA SR results are averaged over three random seeds and reported as mean $\pm$ standard error.
    Methods marked with $^{*}$ could not be reproduced in our setup; their success rates and speedups are quoted from SpecPrune-VLA~\cite{wang2025specprune} (FastV, DivPrune, SparseVLM, EfficientVLA, SpecPrune-VLA) or the original papers (FlashVLA~\cite{tan2025think}, ADP~\cite{pei2025action}), with speedups measured on the authors' hardware, and their FLOPs apply the reported FLOPs ratios to our baseline.}
  \label{tab:libero}
  \centering
  \small
  \setlength{\tabcolsep}{3.3pt}
  \renewcommand{\arraystretch}{1.1}
  \begin{tabular}{lcccccccccc}
    \toprule
    \multirow{2}{*}{Method} &
    \multicolumn{5}{c}{Success Rate (\%) $\uparrow$} &
    \multirow{2}{*}{\makecell{FLOPs\\(T)$\downarrow$}} &
    \multirow{2}{*}{\makecell{Latency\\(ms)$\downarrow$}} &
    \multirow{2}{*}{\makecell{Control Freq.\\(Hz)$\uparrow$}} &
    \multirow{2}{*}{\makecell{CUDA Lat.\\(ms)$\downarrow$}} &
    \multirow{2}{*}{\makecell{Speedup\\$\uparrow$}}\\
    \cmidrule(lr){2-6}
     & Spatial & Object & Goal & 10 & Average & & & & & \\
    \midrule
    OpenVLA & 85.8 & 88.8 & 77.4 & 53.4 & 76.4 & 1.86 & 186.64 & 5.36 & 42.28 & 1.00$\times$ \\
    + VLA-Cache & 86.0 & 88.0 & 75.4 & 54.4 & 76.0 & 1.36 & 203.37 & 4.92 & 31.83 & 0.92$\times$ \\
    + FlashVLA$^{*}$ & 82.6 & 86.6 & 78.8 & 46.8 & 73.7 & 0.82 & - & - & - & 1.56$\times$ \\
    \textbf{+ FastVLA} & \textbf{85.3$\pm$0.1} & \textbf{87.2$\pm$0.2} & \textbf{77.6$\pm$0.2} & \textbf{53.2$\pm$0.4} & \textbf{75.8$\pm$0.2} & \textbf{1.20} & \textbf{161.53} & \textbf{6.19} & \textbf{26.91} & \textbf{1.16$\times$} \\
    \midrule
    OpenVLA-OFT & 98.2 & 98.2 & 97.6 & 94.0 & 97.0 & 3.98 & 118.52 & 67.50 & 65.99 & 1.00$\times$ \\
    + FastV$^{*}$ & 94.6 & 95.8 & 94.0 & 88.8 & 93.3 & 2.27 & - & - & - & 1.44$\times$ \\
    + DivPrune$^{*}$ & 92.4 & 91.2 & 89.0 & 84.8 & 89.4 & 2.15 & - & - & - & 1.46$\times$ \\
    + SparseVLM$^{*}$ & 96.8 & 94.2 & 97.6 & 93.6 & 95.6 & 3.06 & - & - & - & 1.28$\times$ \\
    + VLA-Cache & 98.2 & 98.0 & 97.2 & 94.8 & 97.1 & 3.05 & 99.83 & 80.12 & 62.44 & 1.19$\times$ \\
    + EfficientVLA$^{*}$ & 96.5 & 91.1 & 96.0 & 72.1 & 88.9 & 1.39 & - & - & - & 1.52$\times$ \\
    + SpecPrune-VLA$^{*}$ & 97.4 & 95.8 & 97.7 & 93.4 & 96.1 & 1.71 & - & - & - & 1.46$\times$ \\
    + ADP$^{*}$ & 99.4 & 98.0 & 96.4 & 91.2 & 96.3 & 3.24 & - & - & - & 1.23$\times$ \\
    + LightVLA & 97.6 & 98.0 & 97.4 & 93.0 & 96.5 & 1.21 & 72.41 & 110.48 & 32.67 & 1.64$\times$ \\
    \textbf{+ FastVLA} & \textbf{97.8$\pm$0.1} & \textbf{98.0$\pm$0.1} & \textbf{97.0$\pm$0.5} & \textbf{93.1$\pm$0.5} & \textbf{96.5$\pm$0.3} & \textbf{1.65} & \textbf{81.17} & \textbf{98.56} & \textbf{40.39} & \textbf{1.46$\times$} \\
    \bottomrule
  \end{tabular}
\end{table*}

\section{Experiments}

To evaluate the effectiveness of FastVLA, we conduct experiments in both simulation and real-world environments.
We implement FastVLA on three open-source VLA models: OpenVLA~\cite{kim2024openvla}, OpenVLA-OFT~\cite{kim2025fine}, and CogACT-base~\cite{li2024cogact}.
Specifically, we evaluate OpenVLA and OpenVLA-OFT on the LIBERO benchmark~\cite{liu2023libero} and CogACT-base on the SIMPLER benchmark~\cite{li24simpler}, and additionally validate FastVLA on a physical Franka Research 3 robot, together covering a diverse set of manipulation tasks.
All experiments are conducted on a single NVIDIA RTX 4090 GPU.

\subsection{Experiment Setup}
\textbf{Compared Methods.}
We compare FastVLA against a broad range of recent acceleration methods spanning three categories: (i) general VLM token-pruning methods, including FastV~\cite{chen2024image}, SparseVLM~\cite{zhang2024sparsevlm}, and DivPrune~\cite{alvar2025divprune}; (ii) VLA-specific token compression, caching, and reuse methods, including VLA-Cache~\cite{xu2025vla}, EfficientVLA~\cite{yang2025efficientvla}, FlashVLA~\cite{tan2025think}, SpecPrune-VLA~\cite{wang2025specprune}, and ADP~\cite{pei2025action}; and (iii) the learned token-pruning method LightVLA~\cite{jiang2025better}.
Among these, VLA-Cache and LightVLA provide published, open-source, and reproducible implementations, which we re-evaluate under our unified setup.
For the remaining methods (marked with $^{*}$ in Tab.~\ref{tab:libero}), we are unable to reproduce them in our environment and therefore quote their published results for reference only.

\noindent\textbf{Evaluation Metrics.}
We report success rate, FLOPs, latency, control frequency, and CUDA latency, together with the resulting speedup.
FLOPs provides a hardware-agnostic estimate of the theoretical computation of the LLM backbone.
Latency denotes the total per-step action-prediction time. 
Control frequency is the number of environment actions executed per second, $H/\text{latency}$, where $H$ is the number of actions produced by one prediction call ($H{=}8$ for OpenVLA-OFT, which predicts an action chunk per forward pass, and $H{=}1$ for OpenVLA and CogACT); it quantifies the real-time responsiveness of closed-loop control and makes chunked and single-step policies comparable on one axis. For chunked policies this is an action throughput computed from latency, not a measured rate: OpenVLA-OFT receives a new observation once per chunk, i.e., at most $1/\text{latency}$ ($8.4{\to}12.3$\,Hz in Tab.~\ref{tab:libero}).
CUDA latency reports the LLM-backbone inference time on the GPU, and speedup is computed as the ratio of the per-backbone baseline latency to that of each method.

\noindent\textbf{Choice of $\rho$.} $\rho$ sets FastVLA's average compute budget and is the only setting chosen per benchmark and backbone; $\tau{=}0.1$ and $T{=}10$ are shared by all experiments. For each suite (each task on SIMPLER) we report the most aggressive $\rho$ on a 0.05 grid whose success stayed within 2 points of the uncompressed model.

\subsection{Evaluation Environments}
\textbf{Simulation Benchmarks.} We evaluate on two large-scale simulation benchmarks for language-conditioned robot manipulation. \textbf{LIBERO} comprises four suites (Spatial, Object, Goal, and Long) that test generalization across scene layouts, objects, language-specified goals, and long horizons; we follow the standard protocols of OpenVLA\footnote{https://github.com/openvla/openvla} and OpenVLA-OFT\footnote{https://github.com/moojink/openvla-oft}. \textbf{SIMPLER}, designed to reduce the sim-to-real gap, provides visual matching and variant aggregation settings; we evaluate on Pick Coke Can, Move Near, Open/Close Drawer, and Put Apple in Drawer, following the protocol of CogACT\footnote{https://github.com/microsoft/CogACT} with CogACT-base as the baseline.

\begin{figure*}[t]
    \centering
    \includegraphics[width=\linewidth]{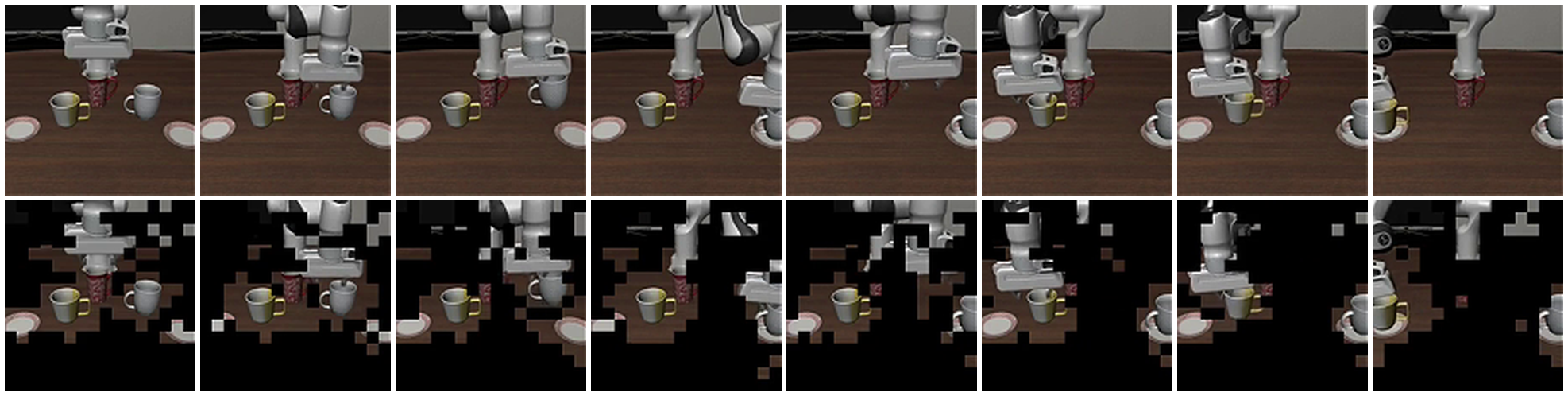}\\[3pt]
    \includegraphics[width=\linewidth]{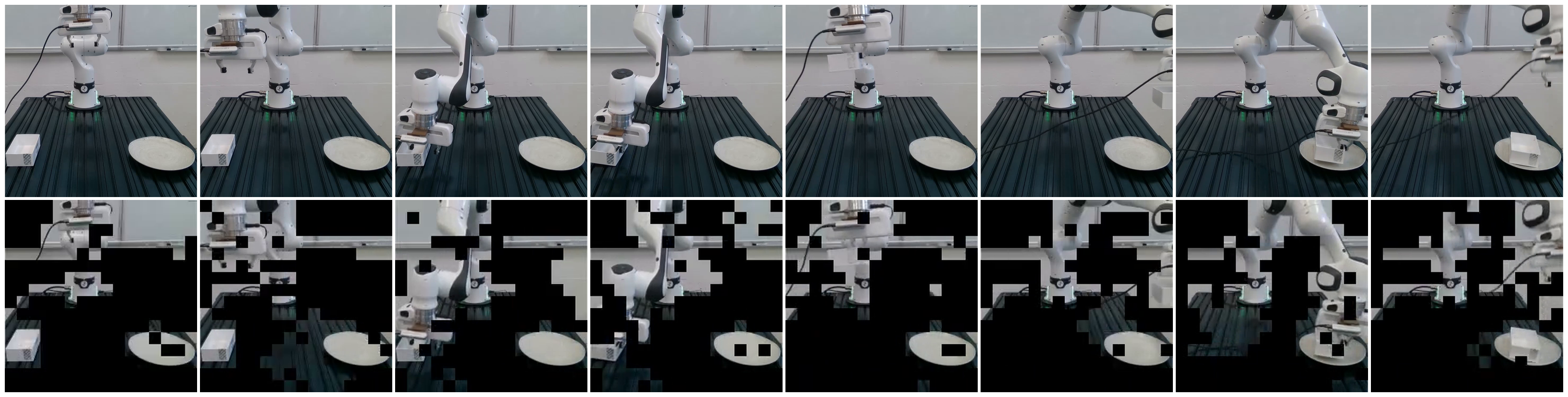}
    \caption{\textbf{Qualitative rollouts.} \emph{Top:} a LIBERO-Long task (simulation); \emph{bottom:} a real-world pick-and-place task. Each strip shows eight frames sampled uniformly over time from the third-person camera, with the full observation alongside its \textit{FastVLA}-pruned counterpart. Full rollout videos are in the supplementary material.}
    \label{fig:rollouts}
\end{figure*}

\noindent\textbf{Real-World Environment.} We further validate FastVLA on a physical Franka Research 3 robot to examine whether the efficiency gains observed in simulation transfer to real deployment.
We consider two tabletop manipulation tasks, \emph{pick} and \emph{pick and place}.
The policy is OpenVLA-OFT fine-tuned with LoRA on teleoperated demonstrations and deployed in closed-loop control using the same RTX 4090 inference setup as in simulation.
For each task and condition, we run 50 rollouts, alternating between the unmodified baseline and FastVLA.

\begin{table*}[!t]
  \caption{Comparison of different VLA acceleration methods on the SIMPLER benchmark. FastVLA SR results are averaged over three random seeds and reported as mean $\pm$ standard error. 
  For FastVLA we set $\rho{=}0.7$ (PickCan), $0.5$ (MoveNear), and $0.3$ (Drawer, DrawerApple).}
  \label{tab:simpler}
  \centering
  \small
  \setlength{\tabcolsep}{4pt}
  \renewcommand{\arraystretch}{1.1}
\resizebox{\textwidth}{!}{%
  \begin{tabular}{llcccccccccc}
    \toprule
    \multirow{2}{*}{SIMPLER} & \multirow{2}{*}{Method} &
    \multicolumn{5}{c}{Success Rate (\%) $\uparrow$} &
    \multirow{2}{*}{\makecell{FLOPs\\(T)$\downarrow$}} &
    \multirow{2}{*}{\makecell{Latency\\(ms)$\downarrow$}} &
    \multirow{2}{*}{\makecell{Control Freq.\\(Hz)$\uparrow$}} &
    \multirow{2}{*}{\makecell{CUDA Lat.\\(ms)$\downarrow$}} &
    \multirow{2}{*}{\makecell{Speedup\\$\uparrow$}}\\
    \cmidrule(lr){3-7}
     & & PickCan & MoveNear & Drawer & DrawerApple & Average & & & & & \\
    \midrule
    \multirow{2}{*}{\makecell[l]{Visual Matching}} & CogACT-base & 93.7 & 86.7 & 75.9 & 53.7 & 77.5 & 1.85 & 174.48 & 5.73 & 82.53 & 1.00$\times$ \\
     & \textbf{+ FastVLA} & \textbf{93.7$\pm$2.0} & \textbf{85.4$\pm$0.7} & \textbf{75.5$\pm$0.9} & \textbf{53.6$\pm$0.2} & \textbf{77.1$\pm$0.8} & \textbf{1.16} & \textbf{159.59} & \textbf{6.27} & \textbf{66.10} & \textbf{1.09$\times$} \\
    \midrule
    \multirow{2}{*}{\makecell[l]{Variant Aggregation}} & CogACT-base & 89.1 & 76.7 & 50.8 & 27.0 & 60.9 & 1.81 & 174.94 & 5.72 & 82.74 & 1.00$\times$ \\
     & \textbf{+ FastVLA} & \textbf{89.5$\pm$1.3} & \textbf{77.2$\pm$1.0} & \textbf{52.4$\pm$0.3} & \textbf{28.3$\pm$0.4} & \textbf{61.9$\pm$0.6} & \textbf{1.13} & \textbf{157.06} & \textbf{6.37} & \textbf{65.34} & \textbf{1.11$\times$} \\
    \bottomrule
  \end{tabular}%
}
\end{table*}

\begin{table}[!t]
  \caption{Real-world evaluation on a Franka Research 3 robot.}
  \label{tab:realworld}
  \centering
  \resizebox{\columnwidth}{!}{%
  \begin{tabular}{lccccc}
    \toprule
    Task & Method &
    \makecell{Success Rate\\ $\uparrow$} &
    \makecell{Latency\\(ms)$\downarrow$} &
    \makecell{Control Freq.\\(Hz)$\uparrow$} &
    \makecell{Speedup\\$\uparrow$} \\
    \midrule
    \multirow{2}{*}{Pick} & OpenVLA-OFT & 84.0\% & 121.13 & 66.04 & 1.0$\times$ \\
     & \textbf{+ FastVLA} & \textbf{80.0\%} & \textbf{96.90} & \textbf{82.56} & \textbf{1.25$\times$}\\
    \midrule
    \multirow{2}{*}{Pick and place} & OpenVLA-OFT & 68.0\% & 120.01 & 66.67 & 1.0$\times$ \\
     & \textbf{+ FastVLA} & \textbf{66.0\%} & \textbf{97.77} & \textbf{81.82} & \textbf{1.23$\times$}\\
    \bottomrule
  \end{tabular}
  }
  \vspace{-4mm}
\end{table}

\begin{table*}[t]
\caption{Compute-matched comparison of fixed and adaptive token selection across four LIBERO suites. 
Tokens (and target compression rate) are reported separately for each suite. Results are averaged over three random seeds and reported as mean $\pm$ standard error.}
\vspace{-2mm}
\centering
\small
\setlength{\tabcolsep}{4pt}
\renewcommand{\arraystretch}{1.1}
\begin{tabular}{lcc|cc|cc|cc}
\toprule
\multirow{2}{*}{Method}
& \multicolumn{2}{c|}{Spatial}
& \multicolumn{2}{c|}{Object}
& \multicolumn{2}{c|}{Goal}
& \multicolumn{2}{c}{10} \\
\cmidrule(lr){2-3}
\cmidrule(lr){4-5}
\cmidrule(lr){6-7}
\cmidrule(lr){8-9}
& Tokens & SR (\%) $\uparrow$
& Tokens & SR (\%) $\uparrow$
& Tokens & SR (\%) $\uparrow$
& Tokens & SR (\%) $\uparrow$ \\
\midrule

Fixed @ nominal
& 102 (80\%) & 96.7$\pm$0.1
& 154 (70\%) & 97.0$\pm$0.2
& 128 (75\%) & 95.1$\pm$0.5
& 205 (60\%) & 91.9$\pm$0.4 \\

Fixed, budget-matched
& 105 & 96.5$\pm$0.1
& 157 & 97.1$\pm$0.1
& 132 & 95.7$\pm$0.3
& 207 & 92.0$\pm$0.3 \\

Adaptive
& 104.3 & \textbf{97.8$\pm$0.1}
& 156.3 & \textbf{98.0$\pm$0.1}
& 131.9 & \textbf{97.0$\pm$0.5}
& 206.8 & \textbf{93.1$\pm$0.5} \\

\bottomrule
\end{tabular}
\label{tab:cr1}
\vspace{-2mm}
\end{table*}

\subsection{Experiment Results}
\textbf{LIBERO.} Tab.~\ref{tab:libero} reports results on the four LIBERO suites. On OpenVLA, FastVLA cuts LLM FLOPs from 1.86 to 1.20\,T ($-35.5\%$) and latency from 186.64 to 161.53\,ms ($1.16\times$) at a cost of 0.6 points of success (76.4$\to$75.8\%); on OpenVLA-OFT, from 3.98 to 1.65\,T ($-58.5\%$) and 118.52 to 81.17\,ms ($1.46\times$, 67.50$\to$98.56\,Hz) at a cost of 0.5 points (97.0$\to$96.5\%). Against prior accelerators the picture is a trade-off rather than dominance. VLA-Cache preserves success best on OpenVLA-OFT (97.1\%) but converts far less computation into speed (FastVLA needs $45.9\%$ fewer FLOPs and $18.7\%$ lower latency), and its gains are backbone-dependent ($0.92\times$ on OpenVLA vs.\ $1.19\times$ on OpenVLA-OFT). Against general VLM pruners (FastV, DivPrune, SparseVLM), FastVLA attains the highest success (96.5\% vs.\ 89--96\%) at lower FLOPs. LightVLA defines the better efficiency frontier: at equal success it uses 26.7\% fewer FLOPs and is 10.8\% faster (1.21 vs.\ 1.65\,T; 72.41 vs.\ 81.17\,ms), but must be re-trained per backbone, whereas FastVLA drops into three backbones and a real robot at inference time with a directly controllable budget $\rho$.

\noindent\textbf{SIMPLER.} Tab.~\ref{tab:simpler} reports results with CogACT-base as the baseline. Under Visual Matching, FastVLA cuts LLM FLOPs by $37.3\%$ and latency by $8.5\%$ ($1.09\times$; $5.73{\to}6.27$\,Hz) with average success essentially unchanged ($-0.4$ points). Under the more challenging Variant Aggregation setting, it cuts FLOPs by $37.6\%$, latency by $10.2\%$ ($1.11\times$; $5.72{\to}6.37$\,Hz) and CUDA latency by $21.0\%$, while improving average success by $1.0$ points, indicating that adaptive compression does not compromise robustness under distribution shift and task variation.

\noindent\textbf{Real-World.} Tab.~\ref{tab:realworld} reports the real-robot results.
On the \emph{pick} task, FastVLA reduces action latency from 121.13 to 96.90 ms ($-20.0\%$) and raises the effective control frequency from 66.04 to 82.56 Hz ($+25.0\%$), while retaining a comparable success rate (84.0\% vs. 80.0\%).
On the more demanding \emph{pick and place} task, latency drops from 120.01 to 97.77 ms ($-18.5\%$) and control frequency increases from 66.67 to 81.82 Hz ($+22.7\%$), with the success rate again largely preserved (68.0\% vs. 66.0\%).
These results confirm that FastVLA's efficiency gains transfer to real-world closed-loop manipulation. Fig.~\ref{fig:rollouts} visualizes the pruning in simulation and on the real robot: the retained tokens concentrate on task-relevant regions such as the robotic arm and the target objects.

\subsection{Ablation Studies}
\label{sec:ablation}

\noindent\textbf{Adaptivity vs.\ the budget it spends.}
Since the step-wise rate lives in a band around $\rho$, an adaptive run may realize a budget slightly different from its nominal target, and a gain could then reflect lighter compression rather than adaptivity. 
We therefore log the \emph{realized} retained-token count and add a compute-matched control: a fixed-rate policy set to the budget the adaptive run actually consumed, $\rho_{\text{fix}}{=}1-\bar{k}/N$ (\cref{tab:cr1}).
Realized budgets exceed the nominal target by $0.9$--$3.0\%$, which is precisely why the control is needed. 
Handing the \emph{fixed} policy that same extra budget is worth at most $+0.6$ points ($+0.15$ on average), whereas the adaptive policy turns it into $+0.9$ to $+1.3$ points on every suite ($95.3{\to}96.5$ on average) while never consuming more tokens than its control. 
The gain thus comes from \emph{when} the budget is spent, not how much of it is spent.

\noindent\textbf{Choice of calibration.}
A rank-based map, $\alpha^{(t)}=\mathrm{rank}(U^{(t)})/T$, is more robust by construction: an anomalous step shifts $\alpha^{(t)}$ by at most $1/T$ instead of resetting the min--max range.
On closed-loop rollouts the two are indistinguishable (\cref{tab:rank_calibration}), differing by at most $0.6$ points ($0.2$ on average) at budgets within $2.5\%$, because $U^{(t)}$ drifts smoothly rather than spiking, so min--max's outlier sensitivity is never exercised. 
We keep min--max, which is continuous and preserves the magnitude of a deviation, and note that rank calibration is a drop-in substitute for noisy observations.

\noindent\textbf{Choice of diversity metric.}
Under identical scoring, budgets and calibration, we compare the inverse-Simpson effective number against a Shannon-based one ($e^{H}/K$) and $1-\text{Gini}$, entropic and order-statistic families, respectively. 
Inverse-Simpson is strongest on both
suites (\cref{tab:diversity_metric}), by $+2.3$ to $+3.0$ points over Shannon and $+1.7$ to $+2.0$ over Gini. 
This is not a budget artifact: it retains only $0.6$--$1.8\%$ more tokens, worth a few tenths of a point by \cref{tab:cr1}. We attribute the advantage to dynamic range: the Shannon logarithm and the saturation of $1-\text{Gini}$ near uniform distributions both compress the variation of $U^{(t)}$, leaving the calibration less contrast to resolve.

\begin{table}[t]
\centering
\caption{Comparison of min--max and rank-based calibration across four LIBERO
suites. Success rates and token usage are reported separately for each suite.}
\small
\setlength{\tabcolsep}{5pt}
\renewcommand{\arraystretch}{1.1}
\begin{tabular}{lcc|cc}
\toprule
\multirow{2}{*}{Calibration}
& \multicolumn{2}{c|}{Spatial}
& \multicolumn{2}{c}{Object} \\
\cmidrule(lr){2-3} \cmidrule(lr){4-5}
& SR (\%) $\uparrow$ & Tokens
& SR (\%) $\uparrow$ & Tokens \\
\midrule
Min--Max & 97.8 & 104.3 & 98.0 & 156.3 \\
Rank     & 97.6 & 103.0 & 98.0 & 159.5 \\
\midrule[\heavyrulewidth]
\multirow{2}{*}{Calibration}
& \multicolumn{2}{c|}{Goal}
& \multicolumn{2}{c}{10} \\
\cmidrule(lr){2-3} \cmidrule(lr){4-5}
& SR (\%) $\uparrow$ & Tokens
& SR (\%) $\uparrow$ & Tokens \\
\midrule
Min--Max & 97.0 & 131.9 & 93.1 & 206.8 \\
Rank     & 96.9 & 132.1 & 92.5 & 205.5 \\
\bottomrule
\end{tabular}
\label{tab:rank_calibration}
\end{table}

\begin{table}[t]
\centering
\caption{Ablation of diversity metrics across the LIBERO-Goal and LIBERO-10 suites. Success rates and token usage are reported separately for each suite.}
\small
\setlength{\tabcolsep}{4pt}
\renewcommand{\arraystretch}{1.1}
\begin{tabular}{lcc|cc}
\toprule
\multirow{2}{*}{Diversity Metric}
& \multicolumn{2}{c|}{Goal}
& \multicolumn{2}{c}{10} \\
\cmidrule(lr){2-3}
\cmidrule(lr){4-5}
& SR (\%) $\uparrow$ & Tokens
& SR (\%) $\uparrow$ & Tokens \\
\midrule

Shannon
& 94.0 & 129.6
& 90.8 & 205.5 \\

Gini
& 95.3 & 130.3
& 91.1 & 205.2 \\

\textbf{Inverse-Simpson}
& \textbf{97.0} & \textbf{131.9}
& \textbf{93.1} & \textbf{206.8} \\

\bottomrule
\end{tabular}

\label{tab:diversity_metric}
\end{table}

\section{Conclusion}
We present FastVLA, a training-free, task-conditioned, and temporally adaptive visual-token compression framework for efficient VLA inference.
Beyond scoring token importance with task-to-vision attention, its core idea is to reformulate per-step token budgeting as an online decision, setting the compression rate from the inverse-Simpson concentration of the importance distribution under sliding-window calibration.
Across three backbones and two simulation benchmarks, FastVLA cuts LLM-backbone FLOPs by 35-59\% and action latency by 9-32\% while maintaining task success, and on a physical Franka robot it reduces latency by 20\% and raises control frequency by up to 25\% at comparable success, staying within 0.6 points of the uncompressed models on LIBERO while remaining entirely training-free. 
Incorporating robot-specific structure such as 3D geometry or proprioception into the importance estimate, and combining adaptive pruning with complementary reuse or early-exit strategies, are promising directions for further accelerating real-time VLA control.

\noindent\textbf{Limitations.} $\rho$ is chosen per benchmark and backbone, and the budget-matched control covers only OpenVLA-OFT on LIBERO. The connection between concentration and safe pruning is supported indirectly by that control rather than by a per-step analysis of pruning-induced action errors. The real-robot study covers two tasks with one backbone, and several baselines are quoted rather than re-run.

{
    \small
    \bibliographystyle{ieeenat_fullname}
    \bibliography{main}
}

\maketitlesupplementary
\appendix
\section{Discussion on FLOPs vs.\ latency.} 

We isolated every stage with CUDA-event timing, for FastVLA and for both re-run baselines (OpenVLA-OFT, 4-suite average; Tab.~\ref{tab:stages}).
Token scoring and selection add negligible cost: FastVLA's scoring takes $1.5$\,ms, $1.8\%$ of a step. 
The gap has three causes. (i)~FLOPs counts only the LLM backbone, which is $55.7\%$ of the baseline step. 
(ii)~Within the LLM stage, time falls sub-linearly in FLOPs because at batch size~1 the backbone is bandwidth- rather than compute-bound at these sequence lengths. (iii)~The vision encoder runs before compression and is unchanged by it; the
residual $5$--$7\%$ spread across methods is run-to-run variation, consistent
with clock throttling under the heavier baseline step. The last row is
computed as the action-prediction latency minus the measured stages, so it absorbs
host--device synchronisation in addition to CPU pre/post-processing; it
shrinks with the token count because the host spends less time waiting on the
backbone and because part of the input preparation scales with sequence
length. 
Crucially, this is a property of the deployment regime and not of FastVLA: all three methods show the same sub-linear behaviour: FLOPs $-58.5\%$ vs.\ LLM time $-38.8\%$ (FastVLA), $-69.6\%$ vs.\ $-50.5\%$ (LightVLA), and $-23.4\%$ vs.\ $-5.5\%$ (VLA-Cache, whose reuse leaves attention spanning the full KV). 
Overall, FastVLA still delivers $118.5{\to}81.2$\,ms ($1.46\times$) and $67.5{\to}98.6$\,Hz.

  \begin{table*}[h]
  \centering\small
  \setlength{\tabcolsep}{3pt}
  \begin{tabular}{lcccc}
  \toprule
  ms/step (4-suite avg.) & Baseline & +FastVLA & +LightVLA & +VLA-Cache \\
  \midrule
  LLM FLOPs (T)$\downarrow$   & 3.98  & 1.65  & 1.21  & 3.05 \\
  \midrule
  Vision encoder              & 34.6  & 32.9  & 32.0  & 32.8 \\
  Token scoring / selection   & --    & 1.5   & 1.3   & CPU$^\dagger$ \\
  LLM backbone (FLOPs scope)  & 66.0  & \textbf{40.4} & 32.7 & 62.4 \\
  Action head                 & 0.5   & 0.5   & 0.6   & 0.5 \\
  Other (CPU + sync)     & 17.4  & 5.9  & 5.8  & 4.1 \\
  \midrule
  Action-prediction latency   & 118.5 & \textbf{81.2} & 72.4 & 99.8 \\
  \bottomrule
  \end{tabular}
  \caption{Stage-wise latency decomposition (CUDA events, RTX 4090). All stages are measured from 500-episode runs. VLA-Cache selects reusable tokens on the CPU, overlapping the GPU stages. }
  \label{tab:stages}
  \end{table*}

\section{Signal transfer to CogACT}
Tab.~\ref{tab:signal_simpler} supports the transferability of our main finding. 
Across all 16 configurations, vision-to-vision scoring consistently performs worst, and its gap from task-conditioned scoring grows with pruning: from $2.7$ points at $30\%$ to $32.2$ points at $90\%$ on average. 
From $30\%$ to $90\%$ pruning, task-conditioned scoring
retains $22$--$57\%$ of its success, compared with only $4$--$15\%$ for vision-only scoring.
This is consistent with our hypothesis: visual saliency captures visual distinctiveness, which need not align with task relevance, whereas task-conditioned scoring can prioritize decision-critical regions as the token budget becomes scarce.

The comparison between task-to-vision and all-to-vision is less consistent. On SIMPLER, they are nearly tied, with task-to-vision leading in 11 of 16 configurations by only $0.8$
points on average. This may reflect SIMPLER's short, template-generated tasks, for which the task tokens provide limited additional information over the full task-aware
sequence. Thus, the robust distinction is task-aware versus task-agnostic
scoring. We use task-to-vision as the default because it performs best on average while using
a smaller query set, making scoring cheaper.

\begin{table}[t]
  \centering
  \small
  \setlength{\tabcolsep}{4pt}
  \begin{tabular}{llcccc}
    \toprule
    \multirow{2}{*}{Setting / Task} & \multirow{2}{*}{Scoring signal}
      & \multicolumn{4}{c}{Pruning rate} \\
    \cmidrule(lr){3-6}
      & & 30\% & 50\% & 70\% & 90\% \\
    \midrule
    \multicolumn{6}{l}{\textit{Visual Matching}} \\
    \midrule
    \multirow{3}{*}{PickCan}
      & Task-to-vision   & 92.7 & 89.3 & 71.7 & 53.0 \\
      & All-to-vision    & 93.7 & 88.7 & 75.0 & 56.7 \\
      & Vision-to-vision & 88.7 & 78.0 & 58.3 & 11.3 \\
    \cmidrule(lr){1-6}
    \multirow{3}{*}{MoveNear}
      & Task-to-vision   & 88.8 & 79.6 & 60.0 & 30.4 \\
      & All-to-vision    & 86.3 & 75.8 & 54.6 & 35.4 \\
      & Vision-to-vision & 85.4 & 73.8 & 46.7 & 12.5 \\
    \midrule
    \multicolumn{6}{l}{\textit{Variant Aggregation}} \\
    \midrule
    \multirow{3}{*}{PickCan}
      & Task-to-vision   & 88.0 & 82.7 & 74.7 & 46.7 \\
      & All-to-vision    & 89.3 & 80.0 & 72.0 & 45.3 \\
      & Vision-to-vision & 86.7 & 74.7 & 56.0 & 5.3 \\
    \cmidrule(lr){1-6}
    \multirow{3}{*}{MoveNear}
      & Task-to-vision   & 76.7 & 75.8 & 49.2 & 22.5 \\
      & All-to-vision    & 75.8 & 73.3 & 46.7 & 20.0 \\
      & Vision-to-vision & 75.0 & 67.5 & 42.5 & 3.3 \\
    \bottomrule
  \end{tabular}
  \caption{Transfer of the importance signal to CogACT on SIMPLER.
  Success rate (\%) under progressive visual-token pruning for the three candidate scoring signals of Sec.~3.2, on both SIMPLER settings.
  }
  \label{tab:signal_simpler}
\end{table}

\section{Ablations on the deviation factor $\tau$ and window length $T$}
We ablate two key hyperparameters of FastVLA on OpenVLA-OFT: the deviation factor $\tau$ and the window length $T$.

\begin{figure}
    \centering
    \includegraphics[width=0.75\linewidth]{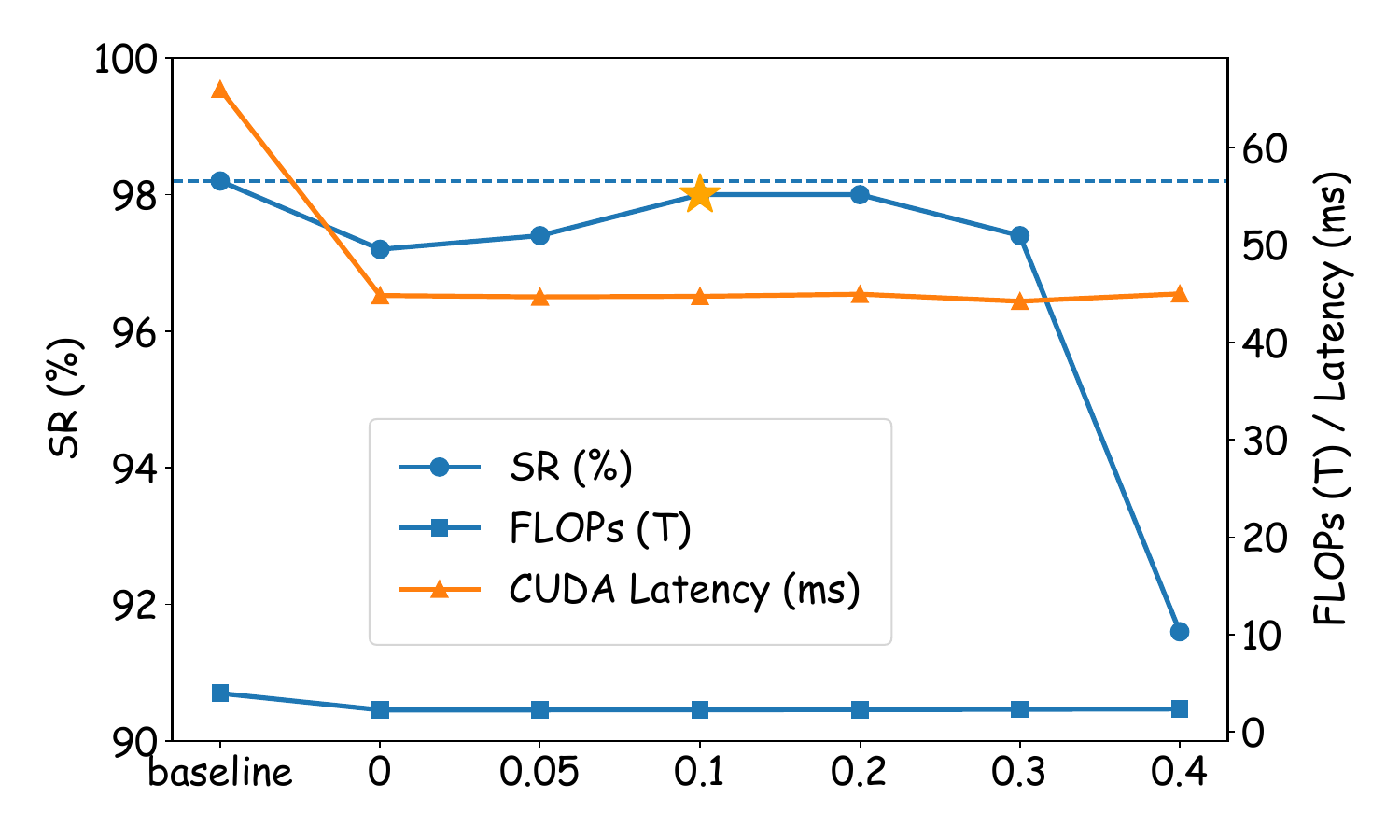}
    \caption{Ablation on the deviation factor $\tau$ in FastVLA on LIBERO-Object using OpenVLA-OFT.}
    \label{tau}
\end{figure}

For the deviation factor $\tau$ (LIBERO-Object, Fig.~\ref{tau}), which controls how aggressively the rate varies across steps, compute is nearly unaffected because it is governed by the target budget while success is highly sensitive: a small $\tau$ ($0$ or $0.05$) reduces to a near fixed-rate strategy and underperforms, an overly large $\tau$ ($\ge0.4$) destabilizes budget allocation and over-compresses at critical moments, and a moderate $\tau$ ($0.1-0.2$) recovers near-baseline success; crucially, the gap between $\tau=0$ and a moderate $\tau$ shows that the temporal adaptivity itself, not merely the average token budget, preserves success, empirically validating the step-adaptive design at the core of FastVLA. 

\begin{figure}
    \centering
    \includegraphics[width=0.75\linewidth]{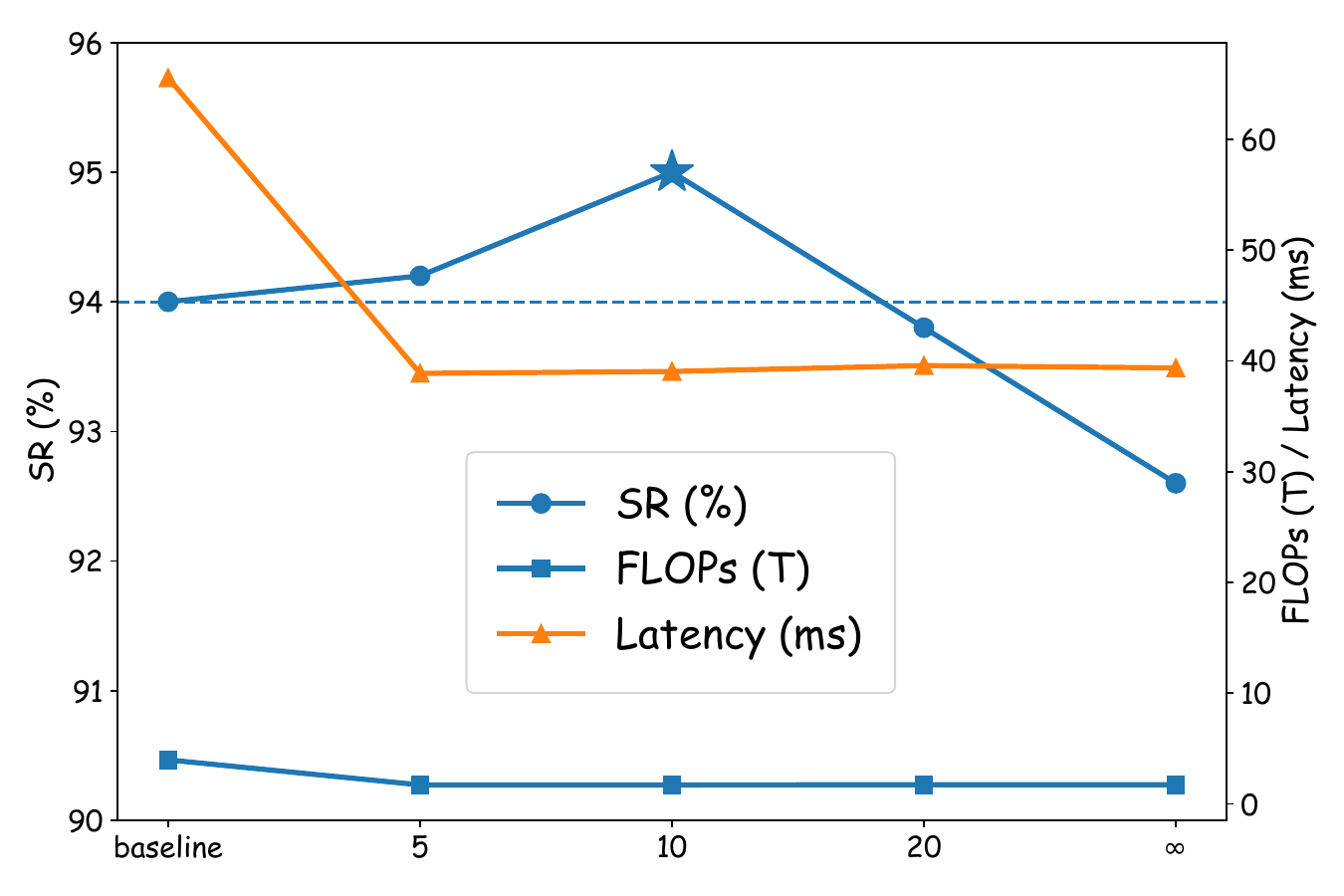}
    \caption{Ablation on the window length $T$ in FastVLA on LIBERO-Long using OpenVLA-OFT.}
    \label{length}
\end{figure}

For the window length $T$ (LIBERO-Long, Fig.~\ref{length}), compute is again negligibly affected while success peaks at a moderate window ($T=10$), outperforming both smaller ($T=5$) and larger ($T=20$ or $+\infty$) windows: too small a window yields unstable temporal statistics, whereas too large a window introduces stale importance signals from earlier manipulation phases that no longer match the current decision.

\end{document}